# Deep Learning-Based Fatigue Cracks Detection in Bridge Girders using Feature Pyramid Networks

Jiawei Zhang[*], Jun Li, Reachsak Ly, Yunyi Liu and Jiangpeng Shu*

*Zhejiang University, Hangzhou, China*

**ABSTRACT**

For structural health monitoring, continuous and automatic crack detection has been a challenging problem. This study is conducted to propose a framework of automatic crack segmentation from high-resolution images containing crack information about steel box girders of bridges. Considering the multi-scale feature of cracks, convolutional neural network architecture of Feature Pyramid Networks (FPN) for crack detection is proposed. As for input, 120 raw images are processed via two approaches (shrinking the size of images and splitting images into sub-images). Then, models with the proposed structure of FPN for crack detection are developed. The result shows all developed models can automatically detect the cracks at the raw images. By shrinking the images, the computation efficiency is improved without decreasing accuracy. Because of the separable characteristic of crack, models using the splitting method provide more accurate crack segmentations than models using the resizing method. Therefore, for high-resolution images, the FPN structure coupled with the splitting method is an promising solution for the crack segmentation and detection.

**INTRODUCTION**

Numerous civil infrastructures such as aged buildings and bridges have gradually approached their design life expectancy, which could cause harmful effects on the safety of people [1]. Therefore, it is necessary to effectively check and evaluate the integrity of the civil structure. However, the currently used human-based visual inspection is limited by the training of the inspector and the high labor cost. There are also problems with unreliable inspection results as well as the time it takes.

As computer vision has made remarkable achievements in various fields, such as object recognition, image segmentation and image classification, more and more researchers start to focus on the research of concrete damage detection based on computer vision. Some of the well-known image processing techniques of computer vision are least square method[2], image binarization methods[3] percolation models[4] and so on. However, most of the traditional image analysis methods focus on crack detection under non-complex conditions. For this reason, its applications are limited because of the great variation of image data with lots of noise and complex background, from engineering practice.

---

[*]Corresponding Author
*Email: jpeshu@zju.edu.cn;  15057126344*

Structural damage inspection is essential for the safety of in-service civil structures, and thus many research groups have utilized the deep learning-based approaches to conduct damage detection on a variety of structures. And because of the great success of deep convolutional neural networks (DCNN), more and more algorithms based on DCNN have been proposed for the application of structural damage inspection like concrete damage detection and crack semantic segmentation. Considering the multilevel and multi-scale features of the crack images, a modified fusion convolutional neural network architecture is proposed[5]. And original crack images are cut into several elements with small size as the input dataset[6]. A three-level deep learning-based method of the inspection of post-disaster bridges was developed with VGG-16[7], Faster R-CNN and SegNet were used to detect system-level failure, component -level and local -level damage respectively. A CNN-based approach was developed to detect bridge damages by acceleration responses[8]. And another five-layer CNN was designed to detect and classify anomalous monitoring data from an SHM system[9]. A semantic segmentation neural network based on SegNet[10] was proposed to automatically localize concrete cracks. Similarly, an U-Net[11] based concrete crack detection framework was developed, which able to identify crack locations under various conditions with the complex background with high efficiency and robustness. Researchers have developed another crack detection system[12] based on FCN, which uses VGG[13] as an encoder. In 2017, Lin et al. [14] introduced a new network architecture called Feature Pyramid Network (FPN). FPN uses a pyramidal hierarchy of deep convolutional networks to construct feature pyramids with marginal extra cost. FPN is renowned for its object detection capabilities. It can improve small object detection by enhancing shallow features. And its architecture shows significant improvement as a generic feature extractor in several applications[15][16][17].

In this paper, to train a model for the semantic segmentation of fatigue crack pixels from given original image dataset, an approach for automatic crack segmentation based on an improved fully convolutional neural network of FPN is proposed. Before inputting the data set into the neural network, two different image preprocessing methods will be performed on the data set, including modification into different picture sizes and separation into different numbers of sub-images. After getting the predicted picture, each mean IoU of the predicted pictures is calculated respectively. In the end, the effectiveness of the proposed methods is discussed and compared.

**DATASET**

There are a total of 200 raw images provided by the International Project Competition organizer. These raw images include crack information inside a steel box girder[6]. Representative raw images are shown in Fig. 1 (a) and (b). Among these 200 raw images, only 120 raw images are annotated in pixel level to indicate whether a pixel is defect-free or belongs to crack damage; as shown in Fig. 1 (c) and (d). Considering that the data set has only 120 images with similar features, it is not suitable for subdividing for competition purposes. Moreover, because there are not many hyperparameters to be set in our model, separating the training set and the test set will not bring better generalization performance. Thus, all 120 annotated raw images are utilized for training and testing to make good use of all annotated raw images. In addition, checking the output of the remaining 80 images without labels could detect whether there is over-fitting (experiments prove that there is no over-fitting phenomenon, indicating that this operation is approximate)

The 120 raw images with annotation and 120 corresponding labels constitute a basic training set. These raw images are in RGB with a resolution of 3264 × 4928 pixels or 3864 × 5152 pixels, and the labels are in grayscale with a resolution of 3264 × 4928 pixels or 3864 × 5152 pixels. It is time-consuming and large memory storage required to directly use images with a high resolution as input. To achieve a trade-off between computation performance and accuracy, two main processes are applied to the images of the basic training set.

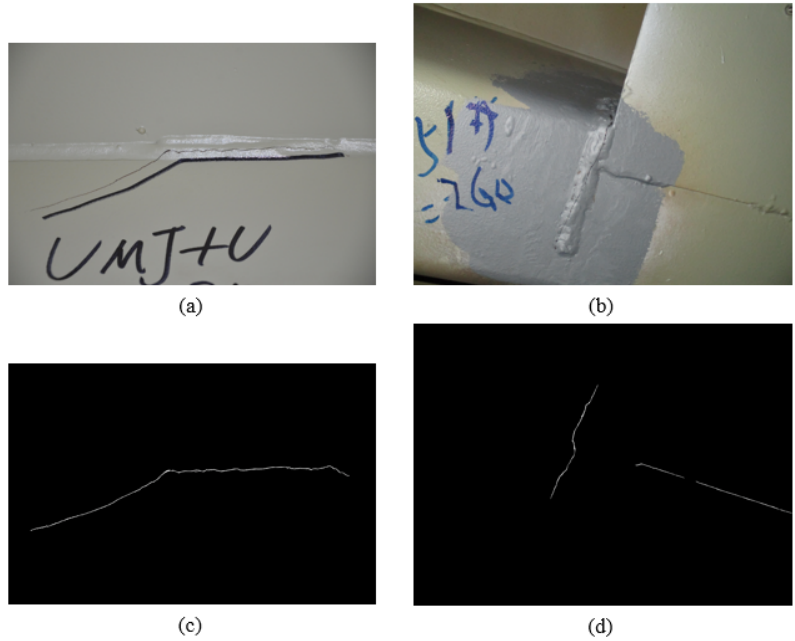

Fig. 1. (a) a raw image of 3264 × 4928 pixels, (b) a raw image of 3864 × 5152 pixels, (c) the label of the raw image (a), (d) the label of the raw image (b).

## THE PROPOSED METHOD

### *Resizing Method*

Images in the basic training set with different resolutions are all resized into 1600 × 2400 pixels or 2112 × 3168 pixels by using a bilinear interpolation method. To examine the influence on predictions caused by the resizing process, two different size changes (1600 × 2400/2112 × 3168) as mentioned above are applied to the images of the basic training set. For convenience, a training set in which images are resized into 1600 × 2400 pixels is named TS1, and a training set in which images are of 2112 × 3168 pixels is denoted as TS2.

*Splitting Method*

Segmentation for cracks doesn't need the full-size object information. Even with the fragment of cracks presented, the crack damage can still be recognized. Thus, another method is proposed to achieve a balance between efficiency and accuracy is splitting images of the basic training set into sub-images. The sub-images are 480 × 640 pixels. Before splitting the images, images of 3264 × 4928/3864 × 5152 pixels are first padding into 3360 × 5120/4320 × 5760 with zeros which can contribute to collecting the edge information of one image. After padding, a stride of 320/320 pixels in rows/columns is applied to the images in the basic training set, getting all 15861 sub-images and 15861 correspond sub-labels. The processing flow is shown in Fig. 2.

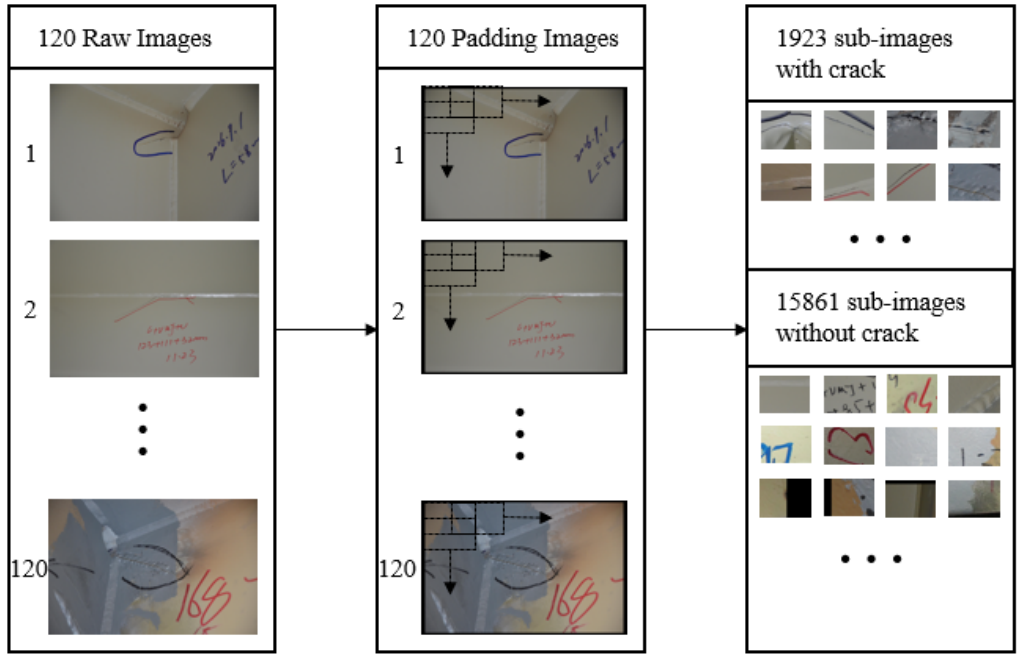

Fig. 2. The splitting process flow in training.

These sub-images are overlapping and some sub-images contain the padding part. Utilizing all 15861 sub-images as input requires much training time. Thus, according to the corresponding sub-labels, 1923 sub-images that contain pixels belonging to crack damage are automatically selected out. Some samples of crack sub-images and corresponding labels are shown in Fig. 3 (a) and (b). These 1923 sub-images make up a training set, named TS3. The rest 13938 sub-images contain no pixel belonging to crack damage and indicate the background information. Some background sub-images are shown in Fig. 3 (c). Among the rest 13938 sub-images, 3500 sub-images are chosen randomly. 1923 crack sub-images and 3500 background sub-images construct another training set, named TS4. When training, image augmentations are often applied to the input images. In Fig. 4, some image augmentation samples are shown.

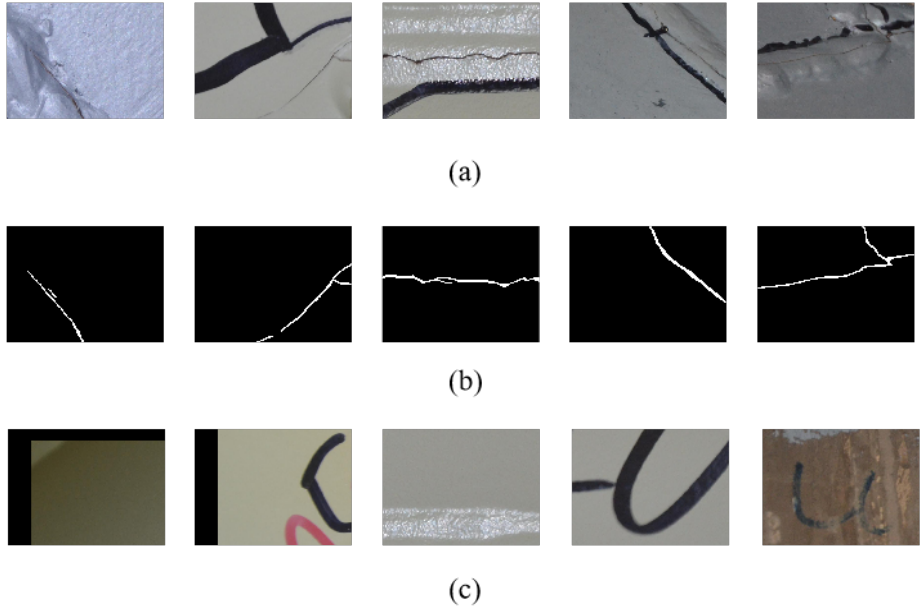

Fig. 3. (a) Sub-images (480×640 pixels) indicating crack damage, (b) Corresponding sub-labels, (c) Sub-images indicating background.

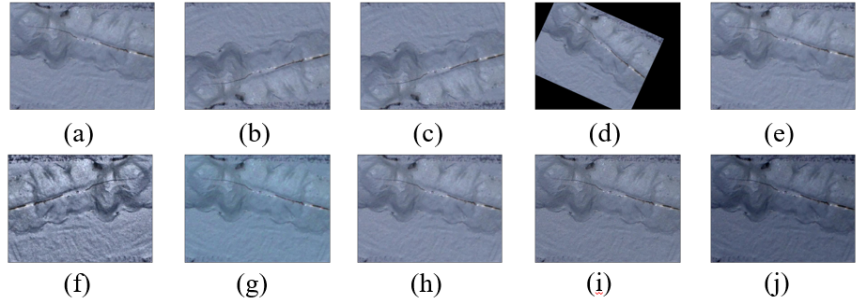

Fig. 4 (a) An original sub-image, and the sub-image through augmentation commands (b) HorizontalFlip, (c) VerticalFlip, (d) ShiftScaleRotate, (e) Blur (blur_limit=3, p=1), (f) CLAHE (p=1), (g) HueSaturationValue (p=1), (h) IAAPerspective (p=0.5), (i) IAASharpen (p=1), (j) RandomBrightness (p=1) in albumentations (an OpenCV library).

## THE PROPOSED METHOD

### The structure of FPN

The network structure is roughly divided into three parts, namely encoder, decoder and assembling, as shown in Fig. 5(a). The encoder is the feedforward process of the convolutional network, which extracts feature vectors at different stages, and the feature resolution is continuously reduced. The decoder is the process of bottom-up feature map enlargement. The top-level feature map is merged with the bottom-level feature map through upsampling to enrich semantic information. After the fusion, a $3 \times 3$ convolution kernel will be used to convolve each fusion result. The purpose is to eliminate the aliasing effect of upsampling.

Finally, in the assembling part, the feature maps of each stage are added to obtain each layer information, which makes the FPN have strong semantic information, and meet the requirements of speed and memory. The difference in FPN is that the prediction is carried out independently in different feature layers, which is helpful to detect crack targets of different sizes.

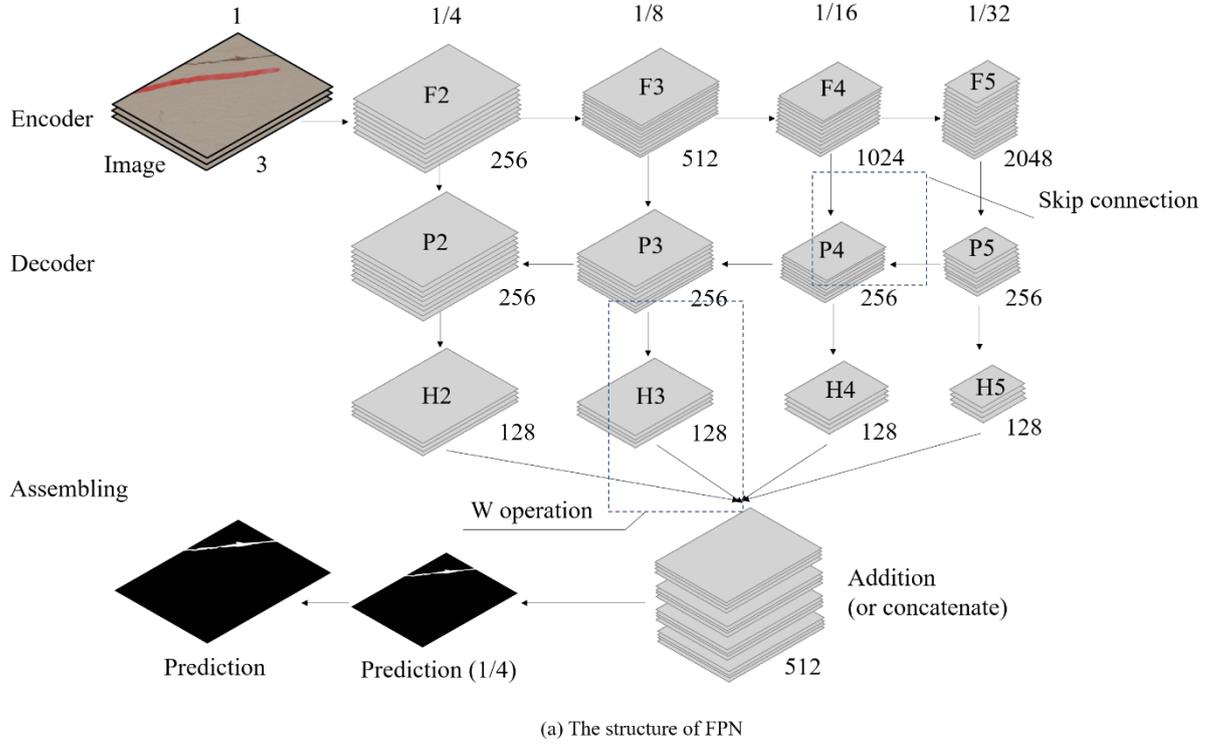

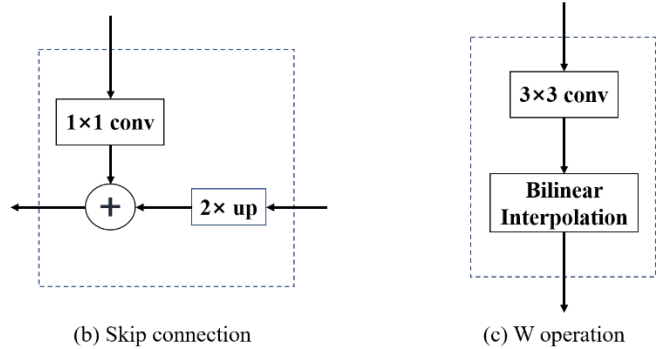

Fig. 5. The structure of FPN: (a) The structure of FPN; (b) Skip connection; (c) W operation.

*Encoder*

The encoder part is a feature extraction network, which generally uses VGG[18] and ResNet[19] as a backbone. To pursue better feature extraction results, se_resnext50_32x4d[20] is adopted. It includes three parts, ResNet, Squeeze-and-Excitation (SE) block, and X. ResNet's residual block allows the network to be deeper, faster to converge, and easier to optimize. At the same time, it has fewer parameters and lower complexity than the previous model, which is suitable for a variety of computer vision tasks. SE block is a computational unit adopted from SENet[20]. Squeeze executes a global average pooling and obtains feature maps with global receptive fields. Excitation uses a fully connected neural network to get a nonlinear transformation on the result after Squeeze, and then use it as a weight to multiply the input feature. SENet mainly learns the correlation between channels, filters out the attention to the channel, and slightly increases the amount of calculation, but the effect is better. X comes from ResNeXt[21], an upgraded version of ResNet. The core innovation of ResNeXt lies in the proposed aggregated transformations, which replaces the original ResNet's three-layer convolutional block with a parallel stacked block of the same topology, which improves the accuracy of the model without significantly increasing the magnitude of the parameter. At the same time, due to the same topology, the hyperparameters are also reduced, which is convenient for model transplantation. ReNext50_32x4d is improved from ResNet50 with 50 layers. 32x4d represents 32 paths, and the number of channels for each path is four. Finally, the SE block is embedded in ReNext50_32x4d to get the final network se_resnext50_32x4d. And its pre-training parameters are obtained from imagenet1000.

Therefore, the encoder part is a bottom-up crack feature extraction network with se_resnext50_32x4d as the backbone. At the entrance of the encoder, the input crack image size is uniformly cropped to 640×480. And a stage corresponds to a level of the feature pyramid in subsequent operations. The features extracted from conv2, conv3, conv4 and conv5 layers are selected as {$C2$, $C3$, $C4$, $C5$}, which are the four levels of FPN network. The feature vectors are $F2 = (N, 256, 120, 160)$, $F3 = (N, 512, 60, 80)$, $F4 = (N, 1024, 30, 40)$, $F5 = (N, 2048, 15, 20)$ respectively. It should be noted that since the last one $F5$ is 1 / 32, the length and width of the crack image should be a multiple of 32.

*Decoder*

The decoder part is a top-down process of enlarging crack feature map. $P5 = (N, 256, 15, 20)$ is directly obtained from $F5$ through a 1×1 convolutional layer. In the following operations, P5 is magnified twice to $(N, 256, 30, 40)$ by interpolation algorithm. $F4$ undergoes a 1×1 convolutional layer and becomes $(N, 256, 30, 40)$. The above two feature vectors are added to form $P4 = (N, 256, 30, 40)$. These operations are called skip connection, as shown in Fig. 5(b), which are continued using to get $P4$, $P3$ and $P2$.

The cleverness of skip connection is that it can make use of both the high-level semantic features of the top layer (helpful for crack feature classification) and the high-resolution information of the bottom layer (helpful for crack feature location).

*Assembling*

For the following assembly assembling operation, the output feature vector of each level of the pyramid should be the same resolution. An operation $W$ contains a 3 × 3 convolution layer and double linear interpolation amplification are chosen to meet the goal, as shown in Fig. 5(c). $P5$ changes to $H5 = (N, 256, 120, 160)$ after three $W$ operations. And so on, $H4$ and $H3$ undergo two and one $W$ operation respectively. Besides, $H2$ does not need to be amplified. Then, $H_i$ $(i = 2\sim5)$ are directly added to get a vector $(N, 256, 120, 160)$. This vector undergoes a 3 × 3 convolution layer and a bilinear interpolation enlarging to original crack image size ($N$, 1, 480, 640). In order to facilitate the prediction process, the mask = ($N$, 1, 480, 640) is obtained by an activation function sigmoid whose values are changed to 0~1. If the value of a point is larger than a fixed threshold, the point is predicted as crack. And in this experiment, 0.5 is chosen as the threshold. According to this mechanism, the criterion of IoU is calculated.

**EXPERIMENT AND DISCUSSION**

All experiments are performed on a computing platform including an Intel(R) Xeon(R) E5-2678 v3 @ 2.50GHz with 64.0GB RAM and an NVIDIA RTX2080TI with 11.0GB RAM. Four models with the same network architecture are trained based on four different training sets. Using the resizing method, Model1 and Model2 are trained with TS1 and TS2 and in the way of the splitting method, Model3 and Model4 are trained with TS3 and TS4. The framework of training is shown in Fig. 6.

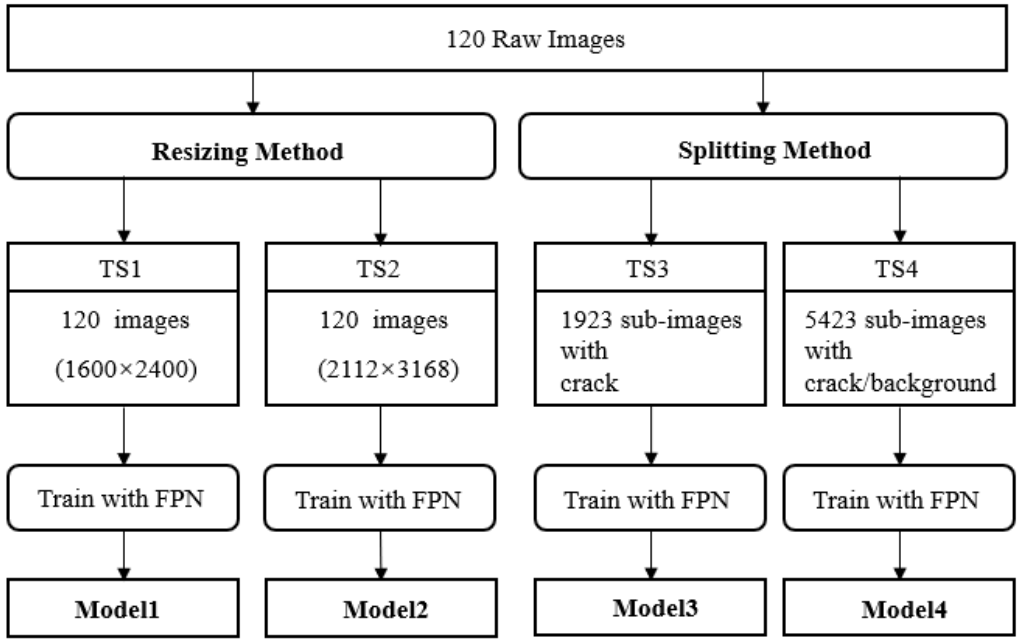

Fig. 6. The framework of training FPN model using 120 raw images and corresponding labels.

According to different trained models, approaches to process a test image are different. Account for the resizing method, outputs of Model1 and Model2 are all resized back to the size of the test image. As for the split method, the test image is first padding with zeros as mentioned in the splitting method above. Then, a stride of 480/640 pixels in rows/columns is applied to the test image to obtain non-overlapping sub-images. To cover the edge of sub-images, using the public point of four adjacent sub-images as the center point, a series of sub-images are obtained again. The processing is shown in Fig. 7. Outputs from Model3 and Model4 are the predictions of sub-images. To get the full-size prediction of the test image, every pixel of outputs then is recombined back to its original position on the test image. The test framework is fully illustrated in Fig. 8.

To validate the performance of trained models, predictions of a test image from the four models are respectively compared with the corresponding label of the test image. Here, two evaluation indexes are used. Dice Loss[22] is a common judgment index of segmentation results. In a crack segmentation task, $X$ is corresponding labels of images; and $Y$ is predictions of images. The Dice Loss is defined as

$$Dice\ Loss = 1 - \frac{2|X \cap Y|}{|X| + |Y|} \tag{1}$$

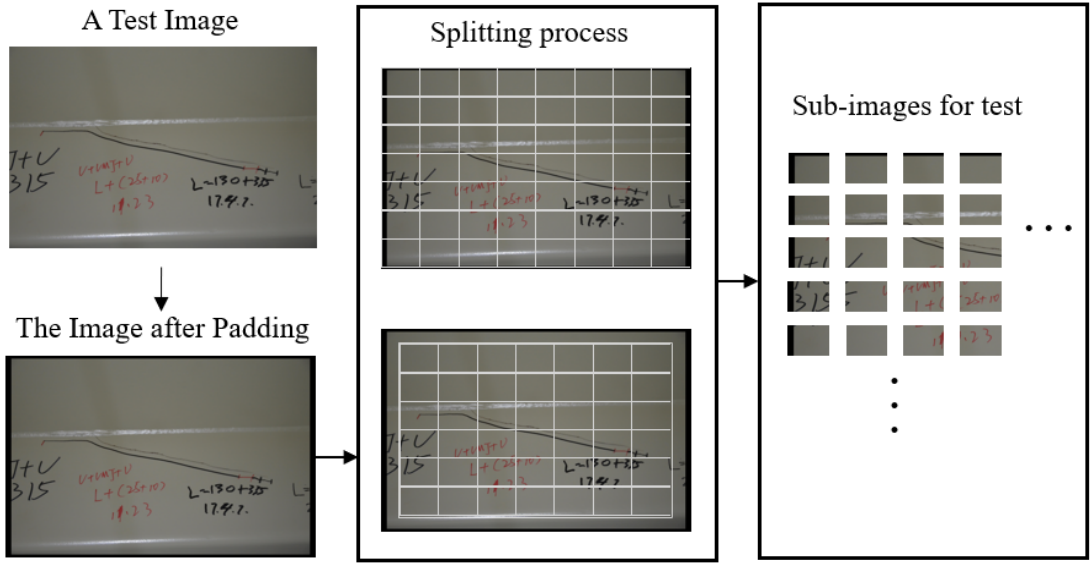

Fig. 7. Splitting process illustration using a test image.

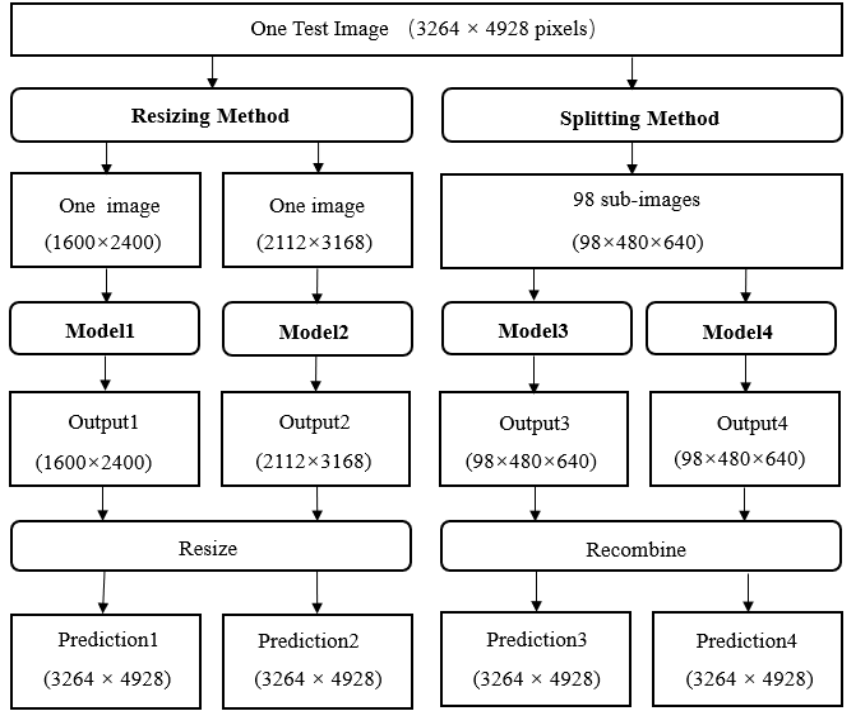

Fig. 8. Framework of the test using trained models with a raw image of 3264 × 4928 pixels.

Another judgment index is IoU (Intersection over Union). The num of pixels which in both label and prediction belong to crack damage is $I$, and the num of pixels which in label or prediction either are indicating crack damage is $U$. The IoU is denoted as

$$IoU = \frac{I}{U} \quad (2)$$

The mIoUs and mean Dice Losses of 120 predictions from four trained models using 120 annotated images as test images are shown in Fig. 9. For convenience, the IoUs of predictions are the main target used to discuss the performance of four models in the following.

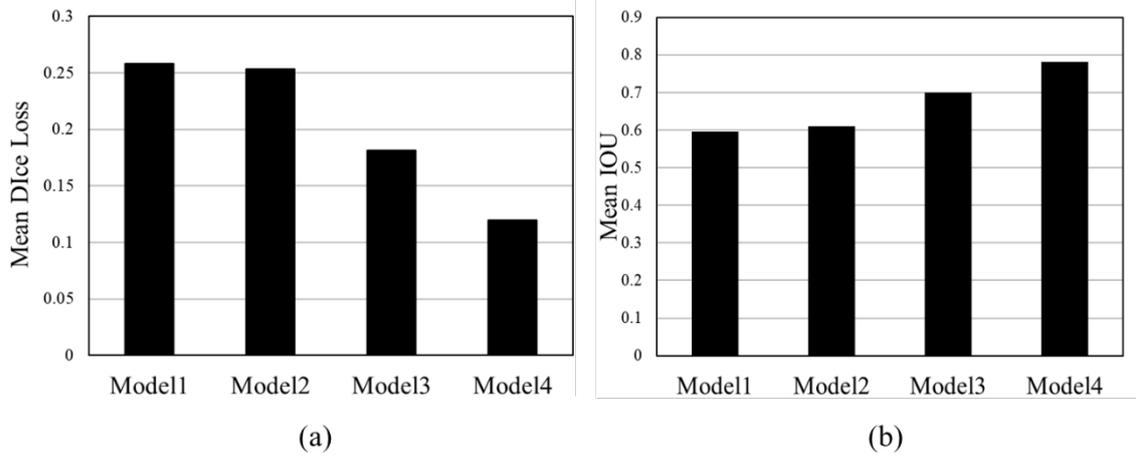

Fig. 9. (a) The mIoUs of four trained models; (b) the mean Dice Losses of four trained models.

From Fig. 9, Model2 shows a bit better performance than Model1 on the 120 test images in terms of IoU and Dice Loss. After analyzing the predictions from Model1 and Model2, one noteworthy phenomenon is observed. As shown in Fig. 10, predictions from Model1 and Model2 both ignore some parts of the cracks in test images, resulting in that the length of cracks in predictions is shorter than cracks in the labels and that some parts of cracks are lost in the predictions. One reason for the phenomenon is the bilinear interpolation algorithm used to resize image shape. The bilinear interpolation considers the closest 2 × 2 neighbourhood of known pixel values surrounding the unknown pixel's computed location. It then takes a weighted average of these 4 pixels to arrive at its final, interpolated value. Using the resizing method means that not every pixel in the test image is predicted, but the statistic values of pixels are predicted. On the one hand, predictions from Model1 and Model2 can't match the label at the pixel level. On the other hand, the predictions can still indicate the positions of the cracks. Based on the positions predicted by Model1 and Model2, some processes can be used to improve the predictions of crack. Crack length can be extended with dilation and threshold algorithm. In Fig. 11(a), the crack in the prediction is extended after processing. But this approach is not suitable for all the predictions. As shown in Fig. 11(b), the black line is considered as crack damage by the Threshold algorithm. With the researcher's manual selection, it may be an acceptable idea to improve the accuracy of Model1 and Model2.

From Fig. 9, the mIoU of Model1 is just 0.02 less than that of Model2. The size of images in TS1 is almost one-third of the image size of TS2. It indicates a much bigger size of images can not improve the predictions too much. However, images of big size as inputs need much memory storage. When resizing the images to get a new training set, computer efficiency, not the size of images, is suggested being first considered.

Without pixels losing, the splitting method achieves improvement on IoUs. The mIoUs of Model3 and Model4 is 0.70 and 0.78. Because Model3 is trained only on the crack sub-images (TS3), some background information is ignored. In Fig. 10(c) and (d), it's shown that Model3 wrongly predicts some pixels on the edge as cracks, and some groove features of the steel box are also predicted as cracks. Although predictions from Model3 are less accurate than those from Model4, Model3 provides a safe estimate in structural health monitoring. From Fig. 10(e), it can be seen that the main crack of the structure is predicted using Model3 and some suspicious features are also considered as cracks, which guarantees no crack damage is missed. Although the mIoU of Model3 is not the best, the Model has an ideal characteristic in engineering. Trained with TS4, Model4 becomes more excellent and more accurate than Model3. Model4 learned both the background information and crack information. In Fig. 10(c), (d), and (e), Model4 has the best performance. The disturbance of background and grooves is dealt with very well. A high degree of accuracy means Model4 can provide precise crack information for the next research or measurement.

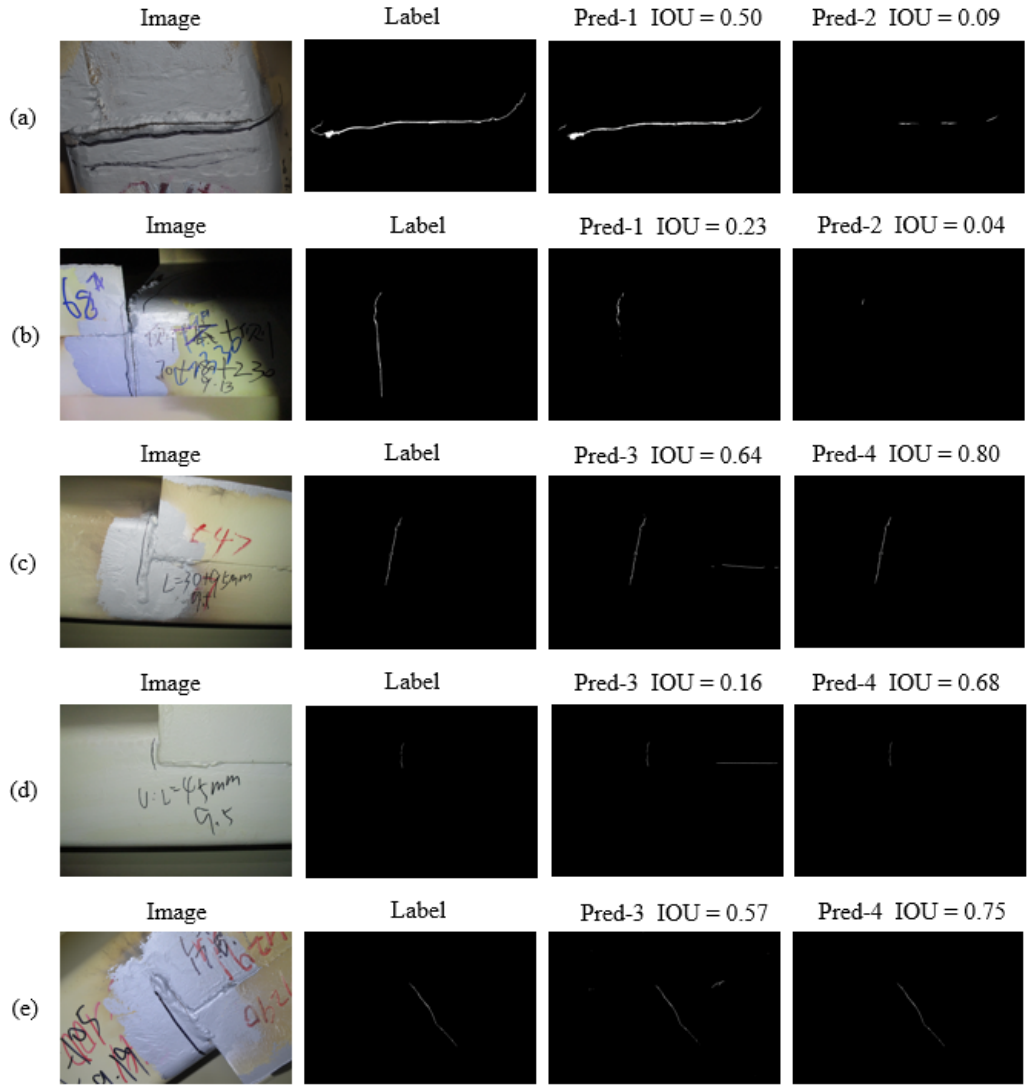

Fig. 10. Some samples of images and predictions (Note that the predictions from Model1 are be abbreviated to Pred-1; Other notations follow the same rule).

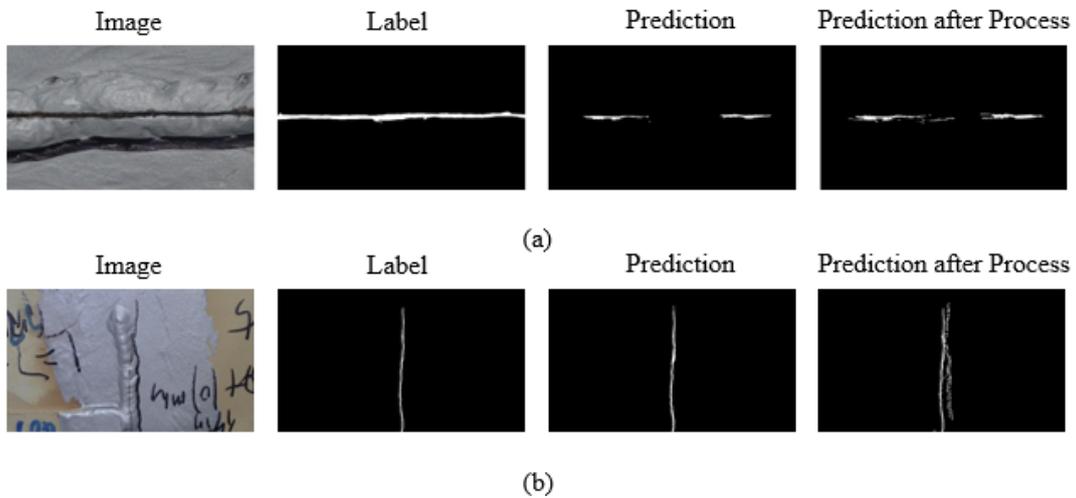

Fig. 11. Samples of processing predictions using the dilate and threshold algorithm.

Model3 and Model4 perform better than Model1 and Model2 for the crack segmentation task. For images of the basic training set, the splitting method is proved as a better solution than the resizing method. Crack segmentation task isn't like the segmentation of cats, dogs, and so on. A part of crack obtains enough information to be detected. Hence, the splitting method is an ideal way of the crack segmentation task. Moreover, the splitting method doesn't have any requirement for the resolution of images. It's suitable for both high-resolution and low-resolution images.

**CONCLUSIONS**

To conduct damage detection using high-resolution image containing both complex backgrounds and cracks inside a steel box girder, an innovative convolutional neural network architecture of Feature Pyramid Networks (FPN) for crack detection is proposed. The performance and accuracy of the proposed framework were validated. Conclusions from the study are summarized as follows:

1) The proposed framework for crack segmentation using FPN coupled with image resizing/splitting method illustrates powerful capability in crack detection based on high-resolution images. The mean IoU reaches 0.78 in maximum. The proposed framework demonstrates a strong sense of semantic information and the ability to detect target cracks of different sizes.

2) For the proposed framework, resizing high-resolution images with a bilinear interpolation algorithm and splitting high-resolution images into sub-images can achieve a balance between computation efficiency and accuracy. When resizing high-resolution images, images can shrink smaller without decreasing segmentation accuracy.

3) Parts of crack can provide enough information for identification. In the segmentation task of crack, splitting images into sub-images is proved as an excellent solution for high-resolution images. By splitting images, every pixel in the images is predicted and pixels belonging to crack damage are precisely annotated. A high degree of accuracy means this method provides a promising solution for the crack detection tasks in the future.

**ACKNOWLEDGEMENTS**

The authors would like to gratefully acknowledge the organizing committee for the platform and elaborate data set. Prof. Jiangpeng Shu is also highly appreciated for the careful guidance of the research and the revision of the paper.